\documentclass{bmvc2k}

\usepackage{enumitem}
\usepackage{amsmath}
\usepackage{amssymb}
\usepackage{amsmath} 
\usepackage{enumitem}
\usepackage{graphicx}
\usepackage{color, colortbl}
\usepackage{tabularx}
\usepackage{tikz}
\usepackage{float}
\usepackage{threeparttable}
\usepackage{subfig}
\DeclareCaptionLabelFormat{singleparen}{#2}
\newcommand{\figref}[1]{Fig.~\ref{#1}}
\newcommand{\tabref}[1]{Table~\ref{#1}}
\newcommand{\eref}[1]{Eq.~(\ref{#1})}
\newcommand{\sref}[1]{Sec.~\ref{#1}}

\newcommand\blfootnote[1]{%
  \begingroup
  \renewcommand\thefootnote{}\footnote{#1}%
  \addtocounter{footnote}{-1}%
  \endgroup
}

\title{BAM: Bottleneck Attention Module}

\addauthor{Jongchan Park\textsuperscript{*$\dagger$}}{jcpark@lunit.io}{1}
\addauthor{Sanghyun Woo\textsuperscript{*}}{shwoo93@kaist.ac.kr}{2}
\addauthor{Joon-Young Lee}{jolee@adobe.com}{3}
\addauthor{In So Kweon}{iskweon77@kaist.ac.kr}{2}

\addinstitution{
 Lunit Inc.\\
 Seoul, Korea
}
\addinstitution{
 Korea Advanced Institute of Science and Technology (KAIST)\\
 Daejeon, Korea
}
\addinstitution{
 Adobe Research\\
 San Jose, CA, USA
}

\runninghead{Park, Woo, Lee, Kweon}{Bottleneck Attention Module}

\def\etal{\emph{et al}\bmvaOneDot}

\begin{document}

\maketitle

\blfootnote{*Both authors have equally contributed.}
\blfootnote{$\dagger$The work was done while the author was at KAIST.}

\begin{abstract}
Recent advances in deep neural networks have been developed via architecture search for stronger representational power. In this work, we focus on the effect of \textit{attention} in general deep neural networks. We propose a simple and effective attention module, named \textit{Bottleneck Attention Module} (BAM), that can be integrated with any feed-forward convolutional neural networks. Our module infers an attention map along two separate pathways, \textit{channel} and \textit{spatial}. We place our module at each \textit{bottleneck} of models where the downsampling of feature maps occurs. Our module constructs a hierarchical attention at bottlenecks with a number of parameters and it is trainable in an end-to-end manner jointly with any feed-forward models. We validate our BAM through extensive experiments on CIFAR-100, ImageNet-1K, VOC 2007 and MS COCO benchmarks. Our experiments show consistent improvement in classification and detection performances with various models, demonstrating the wide applicability of BAM. The code and models will be publicly available\footnote{https://sites.google.com/view/bottleneck-attention-module}.
\end{abstract}

\section{Introduction}

Deep learning has been a powerful tool for a series of pattern recognition applications including classification, detection, segmentation and control problems. Due to its data-driven nature and availability of large scale parallel computing, deep neural networks achieve state-of-the-art results in most areas. Researchers have done many efforts to boost the performance in various ways such as designing optimizers~\cite{zeiler2012adadelta, kingma2014adam}, proposing adversarial training scheme~\cite{goodfellow2014generative}, or task-specific meta architecture like 2-stage architectures~\cite{ren2015faster} for detection.

A fundamental approach to boost performance is to design a good backbone architecture. Since the very first large-scale deep neural network AlexNet~\cite{krizhevsky2012imagenet}, various backbone architectures such as VGGNet~\cite{simonyan2014very}, GoogLeNet~\cite{szegedy2015going}, ResNet~\cite{he2016identity}, DenseNet~\cite{huang2016densely}, have been proposed. All backbone architectures have their own design choices, and show significant performance boosts over the precedent architectures.

The most intuitive way to boost the network performance is to stack more layers. Deep neural networks then are able to approximate high-dimensional function using their \textit{deep} layers. The philosophy of VGGNet~\cite{simonyan2014very} and ResNet~\cite{he2016deep} precisely follows this. Compared to AlexNet~\cite{krizhevsky2012imagenet}, VGGNet has twice more layers. Furthermore, ResNet has 22x more layers than VGGNet with improved gradient flow by adopting residual connections. GoogLeNet~\cite{szegedy2015going}, which is also very deep, uses concatenation of features with various filter sizes at each convolutional block. The use of diverse features at the same layer shows increased performance, resulting in powerful representation. DenseNet~\cite{huang2016densely} also uses the concatenation of diverse feature maps, but the features are from different layers. In other words, outputs of convolutional layers are iteratively concatenated upon the input feature maps. WideResNet~\cite{zagoruyko2016wide} shows that using more channels, \textit{wider} convolutions, can achieve higher performance than naively deepening the networks. Similarly, PyramidNet~\cite{han2017deep} shows that increasing channels in deeper layers can effectively boost the performance. Recent approaches with grouped convolutions, such as ResNeXt~\cite{xie2016aggregated} or Xception~\cite{chollet2016xception}, show state-of-the-art performances as backbone architectures. The success of ResNeXt and Xception comes from the convolutions with higher \textit{cardinality} which can achieve high performance effectively. Besides, a practical line of research is to find mobile-oriented, computationally effective architectures. MobileNet~\cite{howard2017mobilenets}, sharing a similar philosophy with ResNeXt and Xception, use \textit{depthwise convolutions} with high cardinalities.

Apart from the previous approaches, we investigate the effect of \textit{attention} in DNNs, and propose a simple, light-weight module for general DNNs. That is, the proposed module is designed for easy integration with existing CNN architectures.
Attention mechanism in deep neural networks has been investigated in many previous works~\cite{mnih2014recurrent, ba2014multiple, Bahdanau2014, xu2015show, gregor2015draw, Jaderberg2015}.  
While most of the previous works use attention with task-specific purposes, we explicitly investigate the use of attention as a way to improve network's representational power in an extremely efficient way. As a result, we propose ``Bottleneck Attention Module'' (BAM), a simple and efficient attention module that can be used in any CNNs. Given a 3D feature map, BAM produces a 3D attention map to emphasize important elements. In BAM, we decompose the process of inferring a 3D attention map in two streams (\figref{fig:module}), so that the computational and parametric overhead are significantly reduced. As the channels of feature maps can be regarded as feature detectors, the two branches (spatial and channel) explicitly learn `what' and `where' to focus on.

\begin{figure*}[!htb]
\captionsetup{font=footnotesize}
  \centering
  \includegraphics[width=0.9\linewidth]{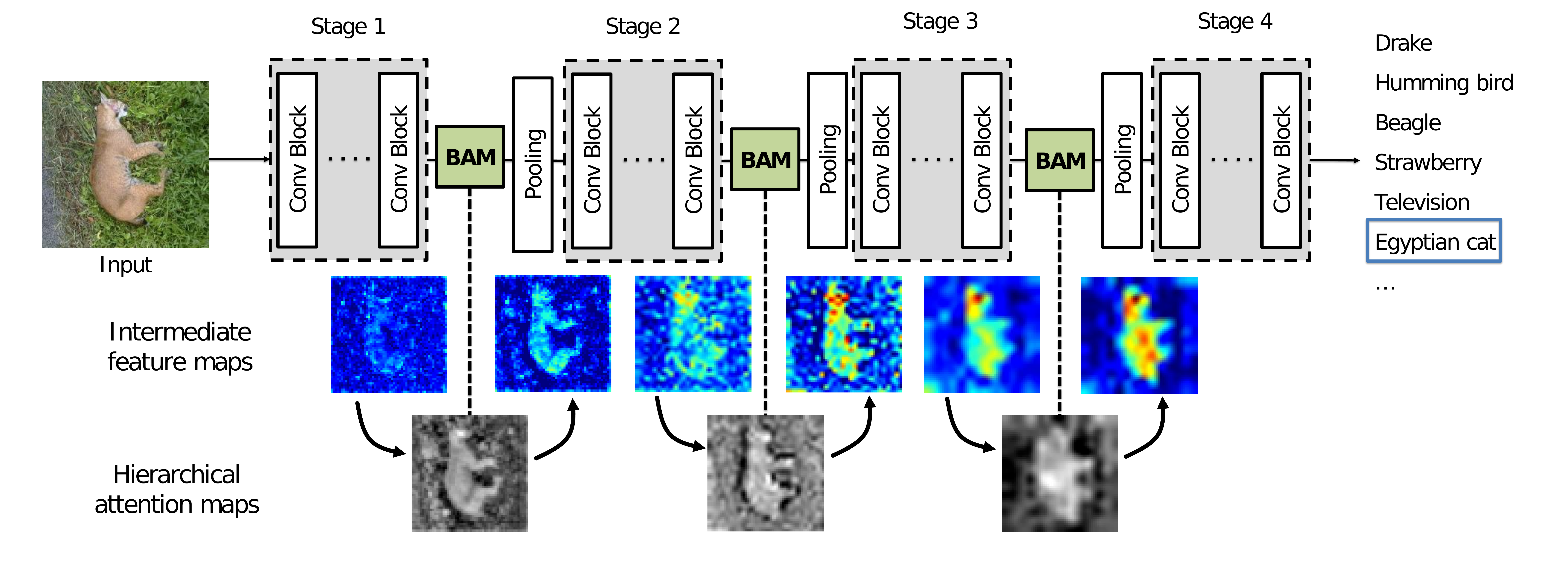}
  \caption{\textbf{BAM integrated with a general CNN architecture.} As illustrated, BAM is placed at every bottleneck of the network. Interestingly, we observe multiple BAMs construct a hierarchical attention which is similar to a human perception procedure. BAM denoises low-level features such as background texture features at the early stage. BAM then gradually focuses on the exact target which is a high-level semantic. \textit{\textbf{More visualizations and analysis}} are included in the supplementary material due to space constraints.}
  \label{fig:teaser}
  \vspace{-5mm}
\end{figure*}

We test the efficacy of BAM with various baseline architectures on various tasks. On the CIFAR-100 and ImageNet classification tasks, we observe performance improvements over baseline networks by placing BAM. Interestingly, we have observed that multiple BAMs located at different bottlenecks build a hierarchical attention as shown in ~\figref{fig:teaser}. Finally, we validate the performance improvement of object detection on the VOC 2007 and MS COCO datasets, demonstrating a wide applicability of BAM. Since we have carefully designed our module to be light-weight, parameter and computational overheads are negligible.

\smallskip\noindent\textbf{Contribution.} Our main contribution is three-fold.
\begin{enumerate}[topsep=0pt,itemsep=0pt]
\item We propose a simple and effective attention module, BAM, which can be integrated with any CNNs without bells and whistles.
\item We validate the design of BAM through extensive ablation studies.
\item We verify the effectiveness of BAM throughout extensive experiments with various baseline architectures on multiple benchmarks (CIFAR-100, ImageNet-1K, VOC 2007 and MS COCO).
\end{enumerate}

\section{Related Work}\label{sec:related}

A number of studies~\cite{Itti1998saliency, rensink2000dynamic, corbetta2002control} have shown that \textit{attention} plays an important role in human perception. For example, the resolution at the foveal center of human eyes is higher than surrounding areas~\cite{hirsch1989spatial}. In order to efficiently and adaptively process visual information, human visual systems iteratively process spatial glimpses and focus on salient areas~\cite{larochelle2010learning}.

\smallskip
\noindent{\textbf{Cross-modal attention.}}
Attention mechanism is a widely-used technique in multi-modal settings, especially where certain modalities should be processed conditioning on other modalities. Visual question answering (VQA) task is a well-known example for such tasks. Given an image and natural language question, the task is to predict an answer such as counting the number, inferring the position or the attributes of the targets. VQA task can be seen as a set of dynamically changing tasks where the provided image should be processed according to the given question. Attention mechanism softly chooses the task(question)-relevant aspects in the image features. As suggested in \cite{yang2016@stacked}, attention maps for the image features are produced from the given question, and it act as \textit{queries} to retrieve question-relevant features. The final answer is classified with the stacked images features. Another way of doing this is to use bi-directional inferring, producing attention maps for both text and images, as suggested in \cite{Nam2017DualAN}. In such literatures, attention maps are used as an effective way to solve tasks in a conditional fashion, but they are trained in separate stages for task-specific purposes.

\smallskip
\noindent{\textbf{Self-attention.}}
There have been various approaches to integrate \textit{attention} in DNNs, jointly training the feature extraction and attention generation in an end-to-end manner. A few attempts~\cite{wang2017residual, hu2017squeeze} have been made to consider \textit{attention} as an effective solution for general classification task. Wang~\etal have proposed \textit{Residual Attention Networks} which use a hour-glass module to generate 3D attention maps for intermediate features. Even the architecture is resistant to noisy labels due to generated attention maps, the computational/parameter overhead is large because of the heavy 3D map generation process. Hu~\etal have proposed a compact `Squeeze-and-Excitation' module to exploit the inter-channel relationships. Although it is not explicitly stated in the paper, it can be regarded as an \textit{attention} mechanism applied upon channel axis. However, they miss the \textit{spatial} axis, which is also an important factor in inferring accurate attention map. 

\smallskip\noindent{\textbf{Adaptive modules.}}
Several previous works use adaptive modules that dynamically changes their output according to their inputs. Dynamic Filter Network~\cite{jia2016dynamic} proposes to generate convolutional features based on the input features for flexibility. Spatial Transformer Network~\cite{jaderberg2015spatial} adaptively generates hyper-parameters of affine transformations using input feature so that target area feature maps are well aligned finally. This can be seen as a \textit{hard attention} upon the feature maps. Deformable Convolutional Network~\cite{dai2017deformable} uses \textit{deformable convolution} where pooling offsets are dynamically generated from input features, so that only the relevant features are pooled for convolutions. Similar to the above approaches, BAM is also a self-contained adaptive module that dynamically suppress or emphasize feature maps through \textit{attention} mechanism.

In this work, we exploit both channel and spatial axes of attention with a simple and light-weight design. Furthermore, we find an efficient location to put our module - \textit{bottleneck} of the network.

\section {Bottleneck Attention Module}

\begin{figure*}[!htb]
\captionsetup{font=footnotesize}
  \centering
  \includegraphics[width=\linewidth]{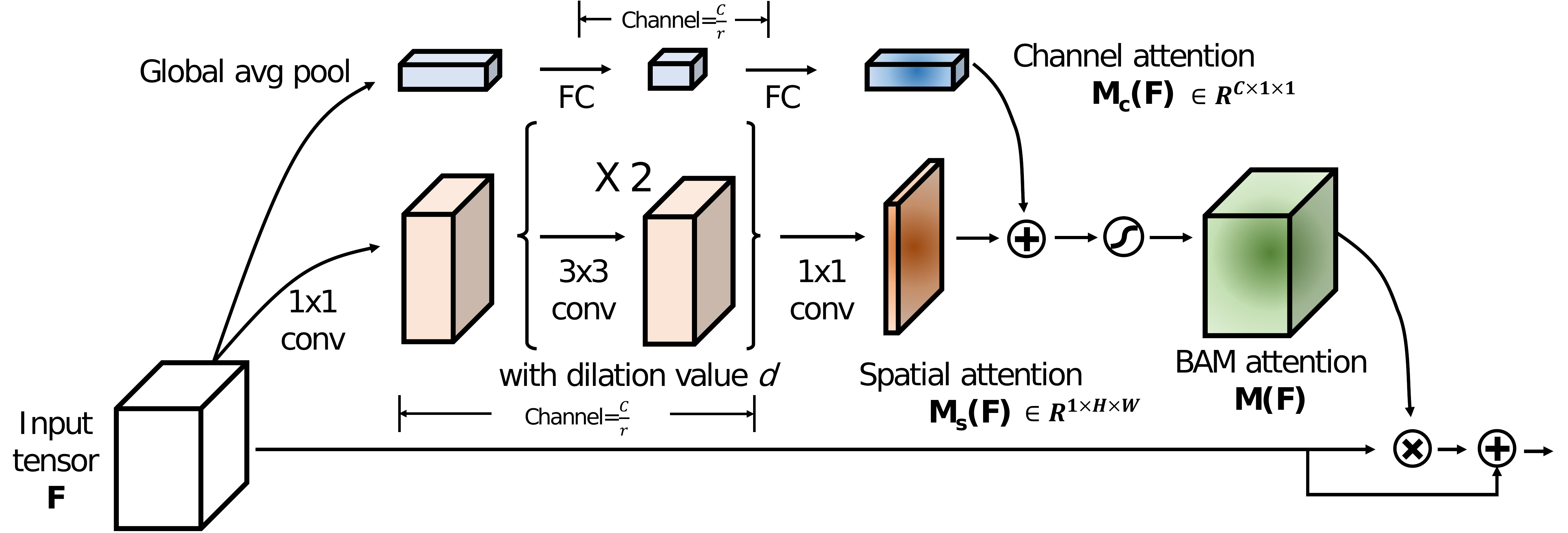}
  \vspace{1mm}
  \caption{\textbf{Detailed module architecture.} Given the intermediate feature map \textbf{F}, the module computes corresponding attention map \textbf{M(F)} through the two separate attention branches -- channel \(\mathbf{M}_\mathbf{c}\) and spatial \(\mathbf{M}_\mathbf{s}\). We have two hyper-parameters for the module: \textit{dilation value (d)} and \textit{reduction ratio (r)}. The dilation value determines the size of receptive fields which is helpful for the contextual information aggregation at the spatial branch. The reduction ratio controls the capacity and overhead in both attention branches. Through the experimental validation (see \sref{sec:ablation}), we set \{d = 4, r = 16\}.}
  \label{fig:module}
  \vspace{-5mm}
\end{figure*}

The detailed structure of BAM is illustrated in ~\figref{fig:module}. For the given input feature map \(\mathbf{F}\in \mathbb{R}^{C\times H\times W}\), BAM infers a 3D attention map \(\mathbf{M}(\mathbf{F})\in \mathbb{R}^{C\times H\times W}\). The refined feature map $\mathbf{F}'$ is computed as:
\begin{equation}\label{eq:first}
    \mathbf{F}'=\mathbf{F}+ \mathbf{F} \otimes \mathbf{M}(\mathbf{F}),
\end{equation}
where $\otimes$ denotes element-wise multiplication.
We adopt a residual learning scheme along with the attention mechanism to facilitate the gradient flow.
To design an efficient yet powerful module, we first compute the channel attention \(\mathbf{M_c}(\mathbf{F})\in \mathbb{R}^{C}\) and the spatial attention \(\mathbf{M_s}(\mathbf{F})\in \mathbb{R}^{H\times W}\) at two separate branches, then compute the attention map \(\mathbf{M}(\mathbf{F})\) as:
\begin{equation}\label{eq:second}
    \mathbf{M}(\mathbf{F})=\sigma(\mathbf{M_c}(\mathbf{F})+\mathbf{M_s}(\mathbf{F})),
\end{equation}
where \(\sigma\) is a sigmoid function. Both branch outputs are resized to \(\mathbb{R}^{C\times H\times W}\) before addition.

\paragraph{Channel attention branch.} 
As each channel contains a specific feature response, we exploit the inter-channel relationship in the channel branch. To aggregate the feature map in each channel, we take global average pooling on the feature map \(\mathbf{F}\) and produce a channel vector \(\mathbf{F_c}\in \mathbb{R}^{C\times 1\times 1}\). 
This vector softly encodes global information in each channel. To estimate attention across channels from the channel vector \(\mathbf{F_c}\), we use a multi-layer perceptron (MLP) with one hidden layer. To save a parameter overhead, the hidden activation size is set to \(\mathbb{R}^{C/r\times 1\times 1}\), where \(r\) is the reduction ratio. After the MLP, we add a batch normalization (BN) layer~\cite{ioffe2015batch} to adjust the scale with the spatial branch output. In short, the channel attention is computed as:
\begin{equation}\label{eq:third}
\begin{split}
    \mathbf{M_c}(\mathbf{F})&=BN(MLP(AvgPool(\mathbf{F})))\\
    &=BN(\mathbf{W_1}(\mathbf{W_0}AvgPool(\mathbf{F})+\mathbf{b_0})+\mathbf{b_1}),
\end{split}
\end{equation}
where \(\mathbf{W_0}\in \mathbb{R}^{C/r\times C}\), \(\mathbf{b_0}\in \mathbb{R}^{C/r}\), \(\mathbf{W_1}\in \mathbb{R}^{C\times C/r}\), \(\mathbf{b_1}\in \mathbb{R}^{C}\).

\paragraph{Spatial attention branch.}
The spatial branch produces a spatial attention map \(\mathbf{M_s}(\mathbf{F})\in \mathbf{R}^{H\times W}\) to emphasize or suppress features in different spatial locations. It is widely known that~\cite{yu2015multi, long2015fcn, bell2016ion, hariharan2015hyper} utilizing contextual information is crucial to know which spatial locations should be focused on. It is important to have a large receptive field to effectively leverage contextual information. We employ the dilated convolution~\cite{yu2015multi} to enlarge the receptive fields with high efficiency. We observe that the dilated convolution facilitates constructing a more effective spatial map than the standard convolution (see \sref{sec:ablation}). The ``bottleneck structure'' suggested by ResNet~\cite{he2016deep} is adopted in our spatial branch, which saves both the number of parameters and computational overhead. Specifically, the feature \(\mathbf{F}\in \mathbb{R}^{C\times H\times W}\) is projected into a reduced dimension \(\mathbb{R}^{C/r\times H\times W}\) using 1$\times$1 convolution to integrate and compress the feature map across the channel dimension. We use the same reduction ratio \(r\) with the channel branch for simplicity. After the reduction, two 3$\times$3 dilated convolutions are applied to utilize contextual information effectively. Finally, the features are again reduced to \(\mathbb{R}^{1\times H\times W}\) spatial attention map using 1$\times$1 convolution. For a scale adjustment, a batch normalization layer is applied at the end of the spatial branch. In short, the spatial attention is computed as:
\begin{equation}\label{eq:forth}
\begin{split}
    \mathbf{M_s}(\mathbf{F})&=BN(f_{3}^{1\times 1}(f_{2}^{3\times 3}(f_{1}^{3\times 3}(f_{0}^{1\times 1}(\mathbf{F}))))),
\end{split}
\end{equation}
where \(f\) denotes a convolution operation, \(BN\) denotes a batch normalization operation, and the superscripts denote the convolutional filter sizes. There are two 1$\times$1 convolutions for channel reduction The intermediate 3$\times$3 dilated convolutions are applied to aggregate contextual information with a larger receptive field.

\paragraph{Combine two attention branches.} 
After acquiring the channel attention \(\mathbf{M_c}(\mathbf{F})\) and the spatial attention \(\mathbf{M_s}(\mathbf{F})\) from two attention branches, we combine them to produce our final 3D attention map \(\mathbf{M}(\mathbf{F})\). Since the two attention maps have different shapes, we expand the attention maps to \(\mathbb{R}^{C\times H\times W}\) before combining them. Among various combining methods, such as element-wise summation, multiplication, or max operation, we choose element-wise summation for efficient gradient flow~\cite{he2016deep}. We empirically verify that element-wise summation results in the best performance among three options (see \sref{sec:experiments}). After the summation, we take a sigmoid function to obtain the final 3D attention map \(\mathbf{M}(\mathbf{F})\) in the range from 0 to 1. This 3D attention map is element-wisely multiplied with the input feature map \(\mathbf{F}\) then is added upon the original input feature map to acquire the refined feature map \(\mathbf{F'}\) as \eref{eq:first}. 

\section {Experiments}\label{sec:experiments}
We evaluate BAM on the standard benchmarks: CIFAR-100, ImageNet-1K for image classification and VOC 2007, MS COCO for object detection. In order to perform better apple-to-apple comparisons, we first reproduce all the reported performance of networks in the PyTorch framework~\cite{pytorch} and set as our baselines~\cite{he2016deep, zagoruyko2016wide, xie2016aggregated, huang2016densely}. Then we perform extensive experiments to thoroughly evaluate the effectiveness of our final module. Finally, we verify that BAM outperforms all the baselines without bells and whistles, demonstrating the general applicability of BAM across different architectures as well as different tasks. Table~5, Table~6, Table~7, Table~8, Table~9 can be found at supplemental material.

\subsection{Ablation studies on CIFAR-100}
\label{sec:ablation}
The CIFAR-100 dataset~\cite{krizhevsky2009learning} consists of 60,000 32$\times$32 color images drawn from 100 classes. The training and test sets contain 50,000 and 10,000 images respectively. We adopt a standard data augmentation method of random cropping with 4-pixel padding and horizontal flipping for this dataset. For pre-processing, we normalize the data using RGB mean values and standard deviations. 

\paragraph{Dilation value and Reduction ratio.}
In \tabref{table:Ablation_1}, we perform an experiment to determine two major hyper-parameters in our module, which are dilation value and reduction ratio, based on the ResNet50 architecture. The \textit{dilation value} determines the sizes of receptive fields in the spatial attention branch. 
\tabref{table:Ablation_1} shows the comparison result of four different dilation values. We can clearly see the performance improvement with larger dilation values, though it is saturated at the dilation value of 4. 
This phenomenon can be interpreted in terms of contextual reasoning, which is widely exploited in dense prediction tasks~\cite{yu2015multi, long2015fcn, bell2016ion, chen2016deeplab, zhu2017couplenet}. 
Since the sequence of dilated convolutions allows an exponential expansion of the receptive field, it enables our module to seamlessly aggregate contextual information. 
Note that the standard convolution (i.e. \textit{dilation value} of 1) produces the lowest accuracy, demonstrating the efficacy of a context-prior for inferring the spatial attention map.
The \textit{reduction ratio} is directly related to the number of channels in both attention branches, which enable us to control the capacity and overhead of our module. In \tabref{table:Ablation_1}, we compare performance with four different reduction ratios. Interestingly, the reduction ratio of 16 achieves the best accuracy, even though the reduction ratios of 4 and 8 have higher capacity. We conjecture this result as over-fitting since the training losses converged in both cases. Based on the result in \tabref{table:Ablation_1}, we set the dilation value as 4 and the reduction ratio as 16 in the following experiments.

\begin{table}
\begin{minipage}{\dimexpr.33\linewidth}
  \centering
  \resizebox{1.0\linewidth}{!}{%
    \begin{tabular}[b]{ l|c|c|c }
        \hline
        Independent Variables & Value & Params & Error \\
        \hline
        \hline
        Dilation value (d)   &1                              &   24.07M  &   20.47\\
                            &2                              &   24.07M  &   20.28\\
                            &\textbf{4}                     &   24.07M  &   \textbf{20}\\
                            &6                              &   24.07M  &   20.08\\
        \hline
        \hline
        Reduction ratio (r)  &4                              &   26.30M  &   20.46 \\
                            &8                              &   24.62M  &   20.56\\
                            &\textbf{16}                    &   24.07M  &   \textbf{20}\\
                            &32                             &   23.87M  &   21.24\\
        \hline
        \hline
        Base (ResNet50\cite{he2016deep})                & -                             &   23.68M  &   21.49  \\
        \hline
    \end{tabular}
  }\par
  {\scriptsize(a) Experiments on hyper-params.}
\end{minipage}%
\begin{minipage}{\dimexpr.35\linewidth}
\centering
\resizebox{1.0\textwidth}{!}{%
    \begin{tabular}[b]{lcccccc}
        \hline
                            &Base    &           &           &             &                & +\textbf{BAM}\\
        \hline
        \hline
            Channel         &         &\checkmark &           &\checkmark   &\checkmark     &\checkmark\\
            Spatial         &         &           &\checkmark &\checkmark   &\checkmark     &\checkmark\\
        \hline
        \hline
            MAX     &         &           &           &\checkmark   &               &           \\
            PROD    &         &           &           &             &\checkmark     &           \\
            SUM     &         &           &           &             &               &\checkmark \\
        \hline
            \textbf{Error}  & 21.49   & 21.29     & 21.24     &20.28        &20.21          &\textbf{20}\\
        \hline
    \end{tabular}%
   }\par
   {\scriptsize(b) Experiments on each branch.}
\end{minipage}
\begin{minipage}{\dimexpr.3\linewidth}
\centering
  \resizebox{1.0\textwidth}{!}{%
    \begin{tabular}[b]{ l|c|c }
        \hline
            ConvBlock vs BAM                & Params             & Error\\
        \hline
        \hline
            ResNet50\cite{he2016deep}                           & 23.68M & 21.49\\
            \quad + ResBlock                                    & 25.14M & 21.02\\
            \quad + \textbf{BAM}                                & \textbf{24.07M} & \textbf{20} \\
        \hline
            WideResNet28\cite{zagoruyko2016wide} (\(w\)=8)      & 23.4M & 20.4\\
            \quad + WideResBlock                                & 24.88M & 19.51 \\
            \quad + \textbf{BAM}                                & \textbf{23.56M}  & \textbf{19.06}\\
        \hline
            ResNeXt\cite{xie2016aggregated} 8x64d               & 34.4M & 18.18 \\
            \quad + ResNeXtBlock                                & 37.3M & 17.69\\
            \quad + \textbf{BAM}                                & \textbf{34.61M} & \textbf{16.71}\\
        \hline
    \end{tabular}%
    }\par
    {\scriptsize(c) Experiments comparing conv blocks and BAM.}
\end{minipage}
\vspace{3mm}
\captionsetup{font=footnotesize}
\caption{Ablation studies on the structure and hyper parameters of BAM. \textbf{(a)} includes experiments for the optimal value for the two hyper parameters; \textbf{(b)} includes experiments to verify the effective of the spatial and channel branches; \textbf{(c)} includes experiments to compare the effectiveness of BAM over the original conv blocks.}
\vspace{-3mm}
    \label{table:Ablation_3}
    \label{table:Ablation_2}
    \label{table:Ablation_1}
\end{table}

\paragraph{Separate or Combined branches.}
In \tabref{table:Ablation_2}, we conduct an ablation study to validate our design choice in the module. We first remove each branch to verify the effectiveness of utilizing both channel and spatial attention branches. As shown in \tabref{table:Ablation_2}, although each attention branch is effective to improve performance over the baseline, we observe significant performance boosting when we use both branches jointly. This shows that combining the channel and spatial branches together play a critical role in inferring the final attention map. In fact, this design follows the similar aspect of a human visual system, which has `what' (channel) and `where' (spatial) pathways and both pathways contribute to process visual information~\cite{larochelle2010learning, chen2016sca}. 

\paragraph{Combining methods.}
We also explore three different combining strategies: element-wise maximum, element-wise product, and element-wise summation. \tabref{table:Ablation_2} summarizes the comparison result for the three different implementations. We empirically confirm that element-wise summation achieves the best performance. In terms of the information flow, the element-wise summation is an effective way to integrate and secure the information from the previous layers. In the forward phase, it enables the network to use the information from two complementary branches, channel and spatial, without losing any of information. In the backward phase, the gradient is distributed equally to all of the inputs, leading to efficient training. Element-wise product, which can assign a large gradient to the small input, makes the network hard to converge, yielding the inferior performance. Element-wise maximum, which routes the gradient only to the higher input, provides a regularization effect to some extent, leading to unstable training since our module has few parameters. Note that all of three different implementations outperform the baselines, showing that utilizing each branch is crucial while the best-combining strategy further boosts performance.

\begin{table}
\centering
\resizebox{0.5\textwidth}{!}{%
\begin{tabular}{ l|c|c|c}
\hline
Architecture            & Params & GFLOPs & Error\\
\hline
\hline
ResNet50\cite{he2016deep}                                         &23.71M  &1.22    &21.49\\
ResNet50\cite{he2016deep} + BAM-C                                  &28.98M  &1.37    &20.88\\
ResNet50\cite{he2016deep} + BAM                                   &\textbf{24.07M}  &\textbf{1.25}    &\textbf{20.00}\\
\hline
\hline
PreResNet110\cite{he2016identity}                                 &1.73M  &0.245   &22.22\\
PreResNet110\cite{he2016identity} + BAM-C                          &\textbf{2.17M}  &\textbf{0.275}        &\textbf{21.29}\\
PreResNet110\cite{he2016identity} + BAM                           &1.73M  &0.246   &21.96\\
\hline
\hline  
WideResNet28 (w=8)\cite{zagoruyko2016wide}                        &23.40M   &3.36   &20.40\\
WideResNet28 (w=8)\cite{zagoruyko2016wide} + BAM-C                 &23.78M   &3.39    &20.06 \\
WideResNet28 (w=8)\cite{zagoruyko2016wide} + BAM                  &\textbf{23.42M}   &\textbf{3.37}
&\textbf{19.06} \\
\hline
\hline
ResNext29 8x64d\cite{xie2016aggregated}                           &34.52M   &4.99    &18.18\\
ResNext29 8x64d\cite{xie2016aggregated} + BAM-C                    &35.60M   &5.07    &18.15\\
ResNext29 8x64d\cite{xie2016aggregated} + BAM                     &\textbf{34.61M}   &\textbf{5.00}    &\textbf{16.71}\\
\hline
\end{tabular}
}
\vspace{3mm}
\captionsetup{font=footnotesize}
\caption{\textbf{Bottleneck v.s. Inside each Convolution Block.} BAM-C denotes where the module is inserted to each convolution block.}
\label{table:conv_vs_bam}
\vspace{-7mm}
\end{table}

\paragraph{Comparison with placing original convblocks.}
In this experiment, we empirically verify that the significant improvement does not come from the increased depth by naively adding the extra layers to the bottlenecks. We add auxiliary convolution blocks which have the same topology with their baseline convolution blocks, then compare it with BAM in \tabref{table:Ablation_3}. we can obviously notice that plugging BAM not only produces superior performance but also puts less overhead than naively placing the extra layers. It implies that the improvement of BAM is not merely due to the increased depth but because of the effective feature refinement.

\paragraph{Bottleneck: The efficient point to place BAM.}
We empirically verify that the bottlenecks of networks are the effective points to place our module BAM. Recent studies on attention mechanisms~\cite{hu2017squeeze, wang2017residual} mainly focus on modifications within the `\textit{convolution blocks}' rather than the `\textit{bottlenecks}'. We compare those two different locations by using various models on CIFAR-100. In \tabref{table:conv_vs_bam}, we can clearly observe that placing the module at the bottleneck is effective in terms of overhead/accuracy trade-offs. It puts much less overheads with better accuracy in most cases except PreResNet 110~\cite{he2016identity}.

\subsection{Classification Results on CIFAR-100}
In \tabref{table:cifar100_exp}, we compare the performance on CIFAR-100 after placing BAM at the bottlenecks of state-of-the-art models including ~\cite{he2016deep, he2016identity, zagoruyko2016wide, xie2016aggregated, huang2016densely}. Note that, while ResNet101 and ResNeXt29~16x64d networks achieve 20.00\% and 17.25\% error respectively, ResNet50 with BAM and ResNeXt29 8x64d with BAM achieve 20.00\% and 16.71\% error respectively using only half of the parameters. It suggests that our module BAM can efficiently raise the capacity of networks with a fewer number of network parameters. Thanks to our light-weight design, the overall parameter and computational overheads are trivial.

\begin{table}
\begin{minipage}{\dimexpr.43\linewidth}
\centering
    \resizebox{1.0\textwidth}{!}{%
        \begin{tabular}[b]{l|l l|c }
        \hline
        Architecture & Parameters & GFLOPs & Error \\
        \hline
        ResNet 50 \cite{he2016deep}                                 & 23.71M                &  1.22        & 21.49\\
        ResNet 50 \cite{he2016deep} + BAM                           & 24.07M$_{(+0.36)}$    &  1.25$_{(+0.03)}$        & \textbf{20.00}\\
        ResNet 101 \cite{he2016deep}                                & 42.70M                &  2.44        & 20.00\\
        ResNet 101 \cite{he2016deep} + BAM                          & 43.06M$_{(+0.36)}$    &  2.46$_{(+0.02)}$        & \textbf{19.61}\\
        \hline\hline
        PreResNet 110\cite{he2016identity}                          &  1.726M               &  0.245             & 22.22\\
        PreResNet 110\cite{he2016identity} + BAM                    &  1.733M$_{(+0.007)}$   &  0.246$_{(+0.01)}$             & \textbf{21.96}\\
        \hline\hline
        WideResNet 28 \cite{zagoruyko2016wide} (\(w\)=8)            & 23.40M                    & 3.36          & 20.40\\
        WideResNet 28 \cite{zagoruyko2016wide} (\(w\)=8) + BAM      & 23.42M$_{(+0.02)}$        & 3.37$_{(+0.01)}$         & \textbf{19.06}\\
        WideResNet 28 \cite{zagoruyko2016wide} (\(w\)=10)           & 36.54M                    & 5.24          & 18.89\\
        WideResNet 28 \cite{zagoruyko2016wide} (\(w\)=10) + BAM     & 36.57M$_{(+0.03)}$        & 5.25$_{(+0.01)}$          & \textbf{18.56}\\
        \hline\hline
        ResNeXt 29 \cite{xie2016aggregated} 8x64d                   & 34.52M                    & 4.99          & 18.18\\
        ResNeXt 29 \cite{xie2016aggregated} 8x64d + BAM             & 34.61M$_{(+0.09)}$        & 5.00$_{(+0.01)}$          & \textbf{16.71}\\
        ResNeXt 29 \cite{xie2016aggregated} 16x64d                  & 68.25M                    & 9.88          & 17.25\\
        ResNeXt 29 \cite{xie2016aggregated} 16x64d + BAM            & 68.34M$_{(+0.09)}$        & 9.90$_{(+0.02)}$          & \textbf{16.39}\\
        \hline\hline
        DenseNet 100-BC\cite{huang2016densely} (\(k\)=12)           & 0.8M                      & 0.29          & 21.95\\
        DenseNet 100-BC\cite{huang2016densely} (\(k\)=12) + BAM     & 0.84M$_{(+0.04)}$       & 0.30$_{(+0.01)}$          & \textbf{20.65}\\
        \hline
        \end{tabular}
  }
  \par
  {\footnotesize(a) CIFAR-100 experiment results}
\end{minipage}
\begin{minipage}{\dimexpr.55\linewidth}
\centering
    \resizebox{1.0\textwidth}{!}{%
        \begin{tabular}[b]{l|ll|c|c }
        \hline
        Architecture  & Parameters & GFLOPs & Top-1(\%) & Top-5(\%) \\
        \hline
        ResNet18 \cite{he2016deep}              & 11.69M & 1.81 & 29.60 &10.55 \\
        ResNet18 \cite{he2016deep} + BAM        & 11.71M$_{(+0.02)}$ & 1.82$_{(+0.01)}$ & \textbf{28.88} &\textbf{10.01}\\
        ResNet50 \cite{he2016deep}              & 25.56M & 3.86 &24.56 & 7.50 \\
        ResNet50 \cite{he2016deep} + BAM        & 25.92M$_{(+0.36)}$ & 3.94$_{(+0.08)}$ &\textbf{24.02} & \textbf{7.18}  \\
        ResNet101 \cite{he2016deep}             & 44.55M & 7.57 & 23.38 & 6.88 \\
        ResNet101 \cite{he2016deep} + BAM       & 44.91M$_{(+0.36)}$ & 7.65$_{(+0.08)}$ &\textbf{22.44} & \textbf{6.29} \\
        
        \hline\hline
        WideResNet18 \cite{zagoruyko2016wide} (widen=1.5)               & 25.88M & 3.87 & 26.85 & 8.88 \\
        WideResNet18 \cite{zagoruyko2016wide} (widen=1.5) + BAM         & 25.93M$_{(+0.05)}$ & 3.88$_{(+0.01)}$ & \textbf{26.67}& \textbf{8.69} \\
        WideResNet18 \cite{zagoruyko2016wide} (widen=2.0)               & 45.62M & 6.70 & 25.63 & 8.20\\
        WideResNet18 \cite{zagoruyko2016wide} (widen=2.0) + BAM         & 45.71M$_{(+0.09)}$ & 6.72$_{(+0.02)}$ & \textbf{25.00} & \textbf{7.81} \\
        \hline\hline
        ResNeXt50 \cite{xie2016aggregated} (32x4d)                      & 25.03M & 3.77 & 22.85  & 6.48 \\
        ResNeXt50 \cite{xie2016aggregated} (32x4d) + BAM                & 25.39M$_{(+0.36)}$ & 3.85$_{(+0.08)}$ & \textbf{22.56} & \textbf{6.40}\\
        \hline
        \hline
        MobileNet\cite{howard2017mobilenets}                                         &4.23M               & 0.569               &31.39            & 11.51\\
        MobileNet\cite{howard2017mobilenets} + BAM                                   &4.32M$_{(+0.09)}$   & 0.589$_{(+0.02)}$   &\textbf{30.58}   & \textbf{10.90}\\
        MobileNet\cite{howard2017mobilenets} \(\alpha=0.7\)                          &2.30M               & 0.283               &34.86            &13.69\\
        MobileNet\cite{howard2017mobilenets} \(\alpha=0.7\) + BAM                    &2.34M$_{(+0.04)}$   & 0.292$_{(+0.009)}$  &\textbf{33.09}   &\textbf{12.69}\\
        MobileNet\cite{howard2017mobilenets} \(\rho=192/224\)                        &4.23M               & 0.439               &32.89            &12.33\\
        MobileNet\cite{howard2017mobilenets} \(\rho=192/224\) + BAM                  &4.32M$_{(+0.09)}$   & 0.456$_{(+0.017)}$  &\textbf{31.56}   &\textbf{11.60}\\
        \hline\hline
        SqueezeNet v1.1~\cite{iandola2016squeezenet}                                 &1.24M               & 0.290               & 43.09           & 20.48\\
        SqueezeNet v1.1~\cite{iandola2016squeezenet} + BAM                           &1.26M$_{(+0.02)}$   & 0.304$_{(+0.014)}$  & \textbf{41.83}  & \textbf{19.58}\\
        \hline
        \end{tabular}
    }\par
    {\footnotesize(b) ImageNet classification results}

\end{minipage}
\begin{tablenotes}
\scriptsize
\item\hspace*{\fill}\textbf{*} all results are reproduced in the PyTorch framework.
\end{tablenotes}
\vspace{3mm}
\captionsetup{font=footnotesize}
\caption{Experiments on image classification tasks: CIFAR-100 and ImageNet 1K classification. The numbers inside the parentheses indicate the parameter/computational overhead. \(w\) denotes the widening factor in WideResNet~\cite{zagoruyko2016wide} and \(k\) denotes the growth rate in DenseNet~\cite{huang2016densely}. For the DenseNet~\cite{huang2016densely}, we put our module back and forth of the transition block.}
\vspace{-5mm}
\label{table:imagenet_exp}
\label{table:cifar100_exp}
\end{table}

\subsection{Classification Results on ImageNet-1K}
The ILSVRC 2012 classification dataset~\cite{deng2009imagenet} consists of 1.2 million images for training and 50,000 for validation with 1,000 object classes. We adopt the same data augmentation scheme with \cite{he2016deep,he2016identity} for training and apply a single-crop evaluation with the size of 224$\times$224 at test time. Following \cite{he2016deep,he2016identity,huang2016deep}, we report classification errors on the validation set. ImageNet classification benchmark is one of the largest and most complex image classification benchmark, and we show the effectiveness of BAM in such a general and complex task. We use the baseline networks of ResNet~\cite{he2016deep}, WideResNet~\cite{zagoruyko2016wide}, and ResNeXt~\cite{xie2016aggregated} which are used for ImageNet classification task. More details are included in the supplementary material.

As shown in \tabref{table:imagenet_exp}, the networks with BAM outperform all the baselines once again, demonstrating that BAM can generalize well on various models in the large-scale dataset. Note that the overhead of parameters and computation is negligible, which suggests that the proposed module BAM can significantly enhance the network capacity efficiently. Another notable thing is that the improved performance comes from placing only three modules overall the network.
Due to space constraints, \textit{\textbf{further analysis and visualizations}} for success and failure cases of BAM are included in the supplementary material.

\subsection{Effectiveness of BAM with Compact Networks}
The main advantage of our module is that it significantly improves performance while putting trivial overheads on the model/computational complexities. To demonstrate the advantage in more practical settings, we incorporate our module with compact networks~\cite{howard2017mobilenets, iandola2016squeezenet}, which have tight resource constraints. Compact networks are designed for mobile and embedded systems, so the design options have computational and parametric limitations.

As shown in \tabref{table:imagenet_exp}, BAM boosts the accuracy of all the models with little overheads. Since we do not adopt any squeezing operation~\cite{howard2017mobilenets, iandola2016squeezenet} on our module, we believe there is more room to be improved in terms of efficiency.

\subsection{MS COCO Object Detection}

We conduct object detection on the Microsoft COCO dataset~\cite{lin2014coco}. 
According to \cite{bell2016ion, liu2016ssd}, we trained our model using all the training images as well as a subset of validation images, holding out 5,000 examples for validation. We adopt \textit{Faster-RCNN}~\cite{ren2015faster} as our detection method and ImageNet pre-trained ResNet101~\cite{he2016deep} as a baseline network. Here we are interested in improving performance by plugging BAM to the baseline. Because we use the same detection method of both models, the gains can only be attributed to our module BAM. As shown in the \tabref{table:coco_detect}, we observe significant improvements from the baseline, demonstrating generalization performance of BAM on other recognition tasks.

\subsection{VOC 2007 Object Detection} \label{sec:voc2007}
We further experiment BAM on the PASCAL VOC 2007 detection task. In this experiment, we apply BAM to the detectors. We adopt the StairNet~\cite{woo2017stairnet} framework, which is one of the strongest multi-scale method based on the SSD~\cite{liu2016ssd}. We place BAM right before every classifier, refining the final features before the prediction, enforcing model to adaptively select only the meaningful features. The experimental results are summarized in \tabref{table:voc_detect}. We can clearly see that BAM improves the accuracy of all strong baselines with two backbone networks. Note that accuracy improvement of BAM comes with a negligible parameter overhead, indicating that enhancement is not due to a naive capacity-increment but because of our effective feature refinement. In addition, the result using the light-weight backbone network~\cite{howard2017mobilenets} again shows that BAM can be an interesting method to low-end devices.

\begin{table}
\begin{minipage}{\dimexpr.5\linewidth}
\centering 
    \resizebox{1.0\textwidth}{!}{%
        \begin{tabular}[b]{ l|c|c|c}
        \hline
        Architecture            & mAP@.5 & mAP@.75 & mAP@[.5,.95] \\
        \hline
        \hline
        ResNet101\cite{he2016deep}      & 48.4           & 30.7          & 29.1     \\
        ResNet101 + BAM                 & \textbf{50.2}  & \textbf{32.5} & \textbf{30.4}     \\
        \hline
        \end{tabular}
    }\par
    {\footnotesize(a) MS-COCO detection task}
\end{minipage}
\begin{minipage}{\dimexpr.5\linewidth}
\centering 
    \resizebox{1.0\textwidth}{!}{%
        \begin{tabular}[b]{ l|l|c|c}
        \hline
        BackBone & Detector & mAP@.5 & params(M) \\
        \hline
        \hline
        VGG16\cite{simonyan2014very} & SSD~\cite{liu2016ssd}        &  77.8  & 26.5\\
        VGG16\cite{simonyan2014very} &
        StairNet~\cite{woo2017stairnet} & 78.9 & 32.0\\
        VGG16\cite{simonyan2014very} &
        \textbf{StairNet~\cite{woo2017stairnet} +
        BAM } & \textbf{79.3} & 32.1\\
        \hline
        \hline
        MobileNet\cite{howard2017mobilenets} & SSD~\cite{liu2016ssd}        &  68.1   & 5.81\\
        MobileNet\cite{howard2017mobilenets} &
        StairNet~\cite{woo2017stairnet} & 70.1 & 5.98\\
        MobileNet\cite{howard2017mobilenets} &
        \textbf{StairNet~\cite{woo2017stairnet} + BAM }                   & \textbf{70.6} &6.00\\
        \hline
        \end{tabular}
    }\par
    {\footnotesize(b) VOC 2007 detection task}
\end{minipage}
\begin{tablenotes}
\scriptsize
\item\hspace*{\fill}\textbf{*} all results are reproduced in the PyTorch framework.
\end{tablenotes}
\vspace{3mm}
\captionsetup{font=footnotesize}
\caption{Experiments on detection tasks: MS-COCO and VOC 2007. Object detection mAP(\%) is reported. We adopt ResNet101-Faster-RCNN as our baseline for MS-COCO, and adopt StairNet~\cite{woo2017stairnet} for VOC 2007.}
\vspace{-2mm}
\label{table:voc_detect}
\label{table:coco_detect}
\end{table}

\vspace{-3mm}

\subsection{Comparison with Squeeze-and-Excitation\cite{hu2017squeeze}}

We conduct additional experiments to compare our method with SE in CIFAR-100 classification task. Table~\ref{table:bam_vs_se} summarizes all the results showing that BAM outperforms SE in most cases with fewer parameters. Our module requires slightly more GFLOPS but has much less parameters than SE, as we place our module only at the bottlenecks not every conv blocks.

\begin{table}
\small
\setlength{\tabcolsep}{3pt}
\begin{center}
\resizebox{0.8\textwidth}{!}{
\begin{tabular}{ l|c|c|c||l|c|c|c}
\hline
Architecture                                    & Params    & GFLOPs    & Error & Architecture                                    & Params    & GFLOPs    & Error\\
\hline
\hline
ResNet50                                        &23.71M     &1.22       &21.49&WideResNet28 (w=8)                              &23.40M     &3.36       &20.40\\
ResNet50 + SE\cite{hu2017squeeze}               &26.24M     &1.23           &20.72&WideResNet28 (w=8) + SE\cite{hu2017squeeze}     &23.58M     &3.36       &19.85 \\
ResNet50 + BAM                                  &24.07M  &1.25 &\textbf{20.00}&WideResNet28 (w=8) + BAM                        &23.42M   &3.37 &\textbf{19.06}\\
\hline
PreResNet110                                    &1.73M      &0.245      &22.22
&ResNext29 16x64d                                &68.25M     &9.88&17.25\\
PreResNet110 + SE\cite{hu2017squeeze}           &1.93M      &0.245           &\textbf{21.85}
&ResNext29 16x64d + SE\cite{hu2017squeeze}       &68.81M     &9.88           &16.52\\
PreResNet110 + BAM                              &1.73M      &0.246      &21.96
&ResNext29 16x64d + BAM                          &68.34M     &9.9&\textbf{16.39}\\
\hline
\end{tabular}
}
\end{center}
\vspace{-4mm}
\begin{tablenotes}
\scriptsize
\item\hspace*{\fill}\textbf{*} all results are reproduced in the PyTorch framework.
\end{tablenotes}
\vspace{2mm}
\captionsetup{font=footnotesize}
\caption{\textbf{BAM v.s. SE~\cite{hu2017squeeze}.} CIFAR-100 experiment results. Top-1 errors are reported.}
\label{table:bam_vs_se}
\vspace{-5mm}
\end{table}

\vspace{-4mm}

\section{Conclusion}
\vspace{-1mm}
We have presented the bottleneck attention module (BAM), a new approach to enhancing the representation power of a network. Our module learns what and where to focus or suppress efficiently through two separate pathways and refines intermediate features effectively. 
Inspired by a human visual system, we suggest placing an attention module at the bottleneck of a network which is the most critical points of information flow.
To verify its efficacy, we conducted extensive experiments with various state-of-the-art models and confirmed that BAM outperforms all the baselines on three different benchmark datasets: CIFAR-100, ImageNet-1K, VOC2007, and MS COCO. In addition, we visualize how the module acts on the intermediate feature maps to get a clearer understanding. Interestingly, we observed hierarchical reasoning process which is similar to human perception procedure. We believe our findings of adaptive feature refinement at the bottleneck is helpful to the other vision tasks as well.

\clearpage
\bibliography{egbib}

\end{document}